\documentclass[letterpaper]{article} 
\usepackage{aaai2026}  
\usepackage{times}  
\usepackage{helvet}  
\usepackage{courier}  
\usepackage[hyphens]{url}  
\usepackage{graphicx} 
\usepackage{natbib}  
\usepackage{caption} 
\usepackage{algorithm}
\usepackage{algorithmic}

\usepackage{newfloat}
\usepackage{listings}
\DeclareCaptionStyle{ruled}{labelfont=normalfont,labelsep=colon,strut=off} 
\floatstyle{ruled}
\newfloat{listing}{tb}{lst}{}
\floatname{listing}{Listing}
\nocopyright

\usepackage{graphicx}
\usepackage{booktabs} 
\usepackage{multirow} 
\usepackage{siunitx} 
\usepackage{caption} 
\usepackage{amsmath}
\title{MMBERT: Scaled Mixture-of-Experts Multimodal BERT for\\Robust Chinese Hate Speech Detection under Cloaking Perturbations}

\author{
Qiyao Xue \quad
Yuchen Dou \quad
Ryan Shi \quad
Xiang Lorraine Li \quad
Wei Gao \quad \\
}
\affiliations {
    \textsuperscript{\rm}University of Pittsburgh\\
    \texttt{\{qix63, yud105, ryanshi, xianglli, weigao\}@pitt.edu}
}

\begin{document}

\maketitle

\begin{abstract}
Hate speech detection on Chinese social networks presents distinct challenges, particularly due to the widespread use of cloaking techniques designed to evade conventional text-based detection systems. Although large language models (LLMs) have recently improved hate speech detection capabilities, the majority of existing work has concentrated on English datasets, with limited attention given to multimodal strategies in the Chinese context. In this study, we propose MMBERT, a novel BERT-based multimodal framework that integrates textual, speech, and visual modalities through a Mixture-of-Experts (MoE) architecture. To address the instability associated with directly integrating MoE into BERT-based models, we develop a progressive three-stage training paradigm. MMBERT incorporates modality-specific experts, a shared self-attention mechanism, and a router-based expert allocation strategy to enhance robustness against adversarial perturbations. Empirical results in several Chinese hate speech datasets show that MMBERT significantly surpasses fine-tuned BERT-based encoder models, fine-tuned LLMs, and LLMs utilizing in-context learning approaches.
\end{abstract}

\section{Introduction}

\begin{figure}[t!]
    \centering
    \includegraphics[width=0.5\textwidth,height=0.45\textwidth]{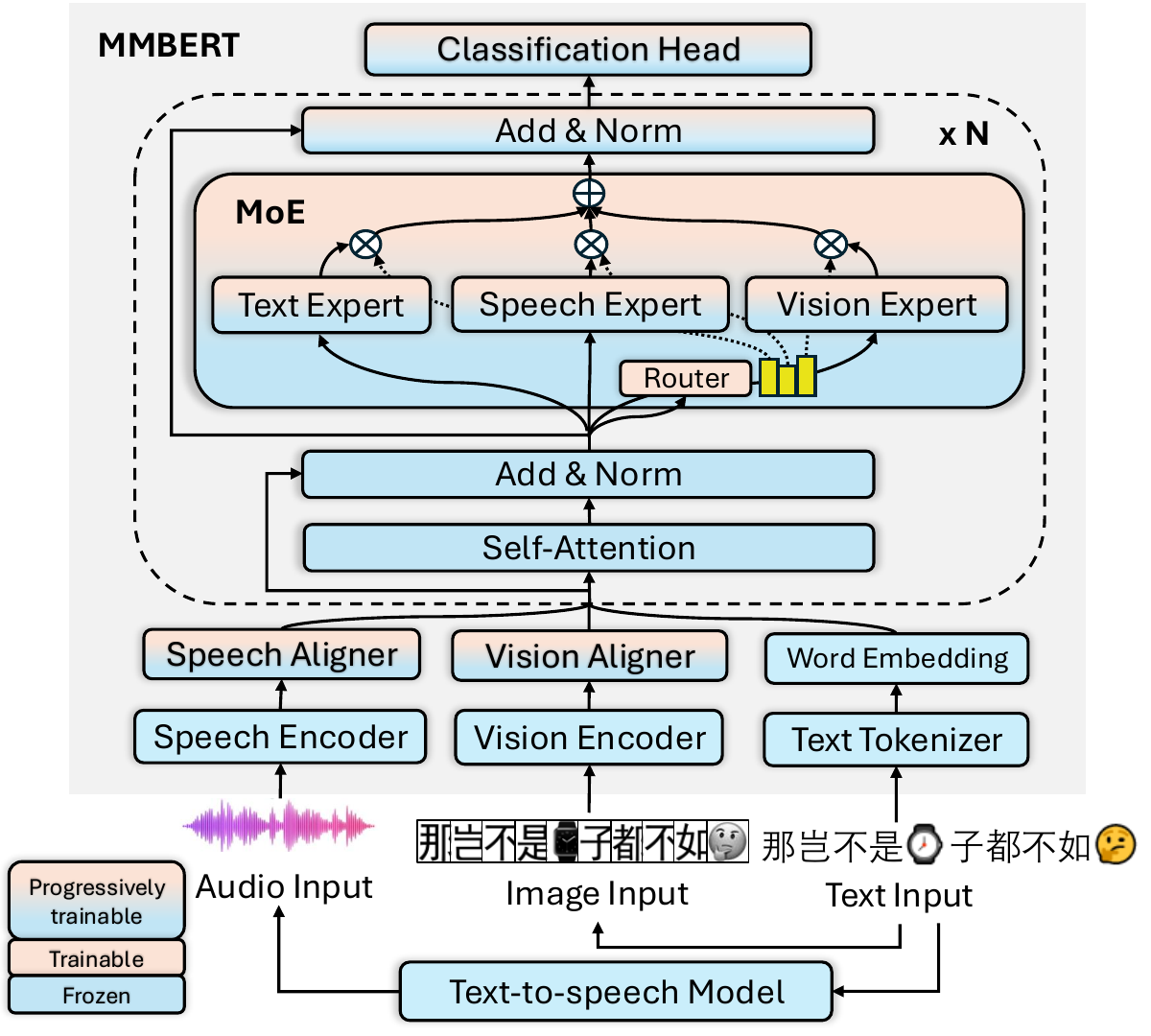}
    \caption{\small \textbf{Illustration of MMBERT model structure}. Compared to traditional BERT-based model, it leverages the MoE architecture to scale and effectively handle multiple modalities. A three-stage progressive training strategy is designed to ensure stable training and prevent performance degradation.} 
    \label{fig:MOE}
\end{figure}

Hate speech poses a persistent threat to online communities, exacerbated by the anonymity and scale of digital platforms \cite{dixon2018measuring}. While automated hate speech detection has advanced significantly in recent years, most efforts remain concentrated on English, leaving other major languages like Chinese relatively  under-resourced and under-protected \cite{davidson2017automated, davidson2019racial}. Some researchers have attempted to leverage LLMs for Chinese hate speech detection \cite{chao2024hate, sun2021chinesebert, zhou2023cross}. However, on Chinese social media platforms, many hate speech disseminators employ various cloaking perturbations to escape detection, making it challenging for existing models to identify such expressions accurately \cite{xiao2024toxicloakcn}. These subtle manipulations exploit the structural and phonological properties of the Chinese language, making detection especially difficult for text-only models.

While LLMs have shown promise in content moderation, BERT-based architectures have consistently outperformed decoder-only LLMs in hate speech detection tasks, owing to their deep bidirectional encoding and strong capacity for fine-grained semantic understanding \cite{benayas2024comparative, ghorbanpour2025can}. Their superior performance can be attributed to the ability to generate fine-grained contextualized representations, which are especially well-suited for classification tasks that require discerning subtle semantic distinctions and interpreting nuanced language—both of which are common in adversarial or implicitly encoded hate speech \cite{liu2024detecting}. The architecture optimized for discriminative tasks enables more efficient and accurate detection of toxic content across various hate speech detection benchmarks \cite{deng2022coldbenchmarkchineseoffensive, xiao2024toxicloakcn}. 

To address the challenge of detecting cloaked hate speech in Chinese, we propose MMBERT, a novel multimodal BERT-based architecture that incorporates visual and speech modalities alongside text, depicted in Figure \ref{fig:MOE}. To enhance scalability and specialization, MMBERT integrates the MoE mechanism, enabling dynamic routing of representations to modality-specific experts. However, naïvely inserting MoE into BERT leads to severe training instability and degraded performance, particularly in the multimodal setting \cite{zhang2021moefication}. To overcome this, we introduce a progressive three-stage training strategy. In the first stage, we pretrain modality aligners using synthetic multimodal data to map visual and auditory inputs into the BERT language space. In the second stage, we train modality-specific experts and continue refining aligners using task-specific supervision. In the final stage, we jointly fine-tune the full MoE-augmented architecture on real multimodal hate speech data. This phased design ensures stable optimization and effective cross-modal integration.

Our experiments across three benchmark Chinese hate speech datasets demonstrate that MMBERT achieves state-of-the-art performance, significantly outperforming both fine-tuned BERT-based baselines and LLMs with in-context learning. In particular, MMBERT shows superior robustness in detecting cloaked adversarial content, highlighting the value of multimodal modeling and progressive training for Chinese hate speech detection.

We summarize the main contribution of this paper as follows:

\begin{itemize}
\item We propose \textbf{MMBERT}, a novel multimodal BERT-based framework for Chinese hate speech detection that integrates textual, visual, and speech modalities through a Mixture-of-Experts (MoE) architecture, enhancing robustness against cloaking-based adversarial perturbations.
\item We design a \textbf{progressive three-stage training strategy} that first aligns multimodal inputs to the BERT language space, then specializes modality-specific experts, and finally fine-tunes the complete model. This approach ensures stable training and effective cross-modal representation learning.
\item We conduct \textbf{extensive experiments} on three benchmark datasets, comparing MMBERT against fine-tuned BERT-based and open-source LLM baselines and closed-source LLMs with in-context learning. Results demonstrate that MMBERT consistently achieves superior performance, particularly in detecting cloaking perturbed hate speech.
\end{itemize}

\section{Background and Motivation}
\subsection{Cloaking Perturbations in Chinese Hate Speech}

Cloaking perturbations in Chinese online discourse represent a growing challenge for automated hate speech detection systems, as users employ various strategies to obfuscate offensive content while preserving its intended meaning \cite{xiao2024toxicloakcn, xiao2024chinese}. It can be mainly categorized into several types:

\textbf{Deformation}. As Chinese characters are logographic, their meanings can be altered by decomposing or reconfiguring individual components, often imparting specific emotional or ideological connotations \citep{lan2006introduction}. For example, the character ``\includegraphics[width=1em]{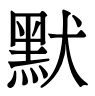}" (meaning `silence') comprises the radicals  ``\includegraphics[width=1em]{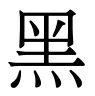}" (meaning `black') and  ``\includegraphics[width=1em]{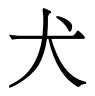}'' (meaning `dog'), which in certain contexts have been used to convey derogatory implications toward the Black community.

\textbf{Homophonic Substitution}. Words with similar pronunciations are frequently substituted to generate alternative semantics \citep{tien2021anatomy}. For instance, Chinese internet users often replace the character ``\includegraphics[width=1em]{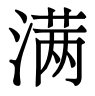}'' (meaning `full') with ``\includegraphics[width=1em]{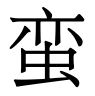}'' (meaning `barbarian'), as both share a phonetic resemblance to `man'.

\textbf{Abbreviation}. The contraction of sensitive terms enhances conciseness while maintaining semantic clarity \citep{lan2006introduction}. A notable example is `txl', where each letter corresponds to the pinyin initials of ``\includegraphics[width=1em]{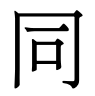}''  ``\includegraphics[width=1em]{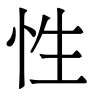}''  ``\includegraphics[width=1em]{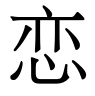}'', collectively denoting `homosexuality'.

\textbf{Code-Mixing}. To intensify expressive tone and circumvent automated content moderation, Chinese social media users frequently incorporate non-Chinese linguistic elements such as pinyin and emojis \citep{li2020swearwords}. These code-mixed constructs not only obscure semantic intent from detection systems but also reinforce the emotive or derogatory force of the message. For instance, the term  ``\includegraphics[width=1.7em]{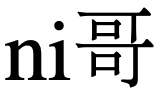}'' (meaning `ni brother') phonetically approximates the English racial slur `n*gger'. Similarly, in the phrase ``\includegraphics[width=1.8em]{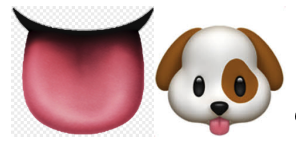}'' (meaning `licking dog'), the addition of an emoji amplifies the pejorative undertone, characterizing individuals perceived as excessively submissive in relationship contexts—analogous to the English term `sycophant'.

These perturbations exploit the unique structural and phonological characteristics of the Chinese language to conceal offensive intent \cite{lu2023facilitating}. For instance, visually altering character radicals can introduce ideological connotations, while homophones and abbreviations obscure meanings through phonetic similarity or reduction. Code-mixing with pinyin or emojis further complicates semantic interpretation. Text-only models often fail to capture these manipulations due to their limited capacity to disambiguate subtle visual and phonological cues \cite{xiao2024chinese, raza2025detecting}. 

\subsection{Enhancing Chinese Language Modeling through Multimodal Pretraining}

Text-only approaches in Chinese language modeling often face limitations in capturing the full linguistic complexity of the language, particularly with respect to character homographs and tonal ambiguity. These challenges hinder the model’s ability to accurately interpret semantic and phonetic nuances inherent in Chinese.

To address these limitations, several studies have explored the integration of additional modalities, such as visual and phonetic information, into the pretraining process. For instance, ChineseBERT \cite{sun2021chinesebert} integrates both glyph and pinyin embeddings, enriching the representation of Chinese characters by capturing visual features through multiple font variations and phonetic information to resolve the heteronym phenomenon. This dual-embedding approach has shown significant improvements in various Chinese natural language processing tasks, such as named entity recognition and sentiment analysis. Similarly, models like ERNIE-M \cite{ouyang2020ernie} and GlyphBERT \cite{li2021glyphcrm} have demonstrated the benefits of incorporating external modalities, such as entity knowledge and visual cues, to enhance language understanding. 

However, existing multimodal approaches predominantly rely on embedding-level fusion of heterogeneous input modalities within a fixed BERT encoder architecture. While such integration enhances input representations, the processing and interaction of multimodal information remain largely static and inflexible. Specifically, the fixed fusion mechanism in standard BERT layers may limit the model’s capacity to dynamically adapt to context-dependent linguistic challenges, such as homographs and tonal ambiguity in Chinese. This rigidity restricts the model's ability to effectively leverage the complementary strengths of each modality in a nuanced and input-sensitive manner.

\subsection{Scaling Multimodal Language Models with MoE Architectures}

Recent advancements in large MLLMs have increasingly explored the use of MoE \cite{eigen2013learning} architectures to enhance scalability, efficiency, and specialization across modalities. Early generations of MLLMs, such as Flamingo \cite{alayrac2022flamingo} and GPT-4V \cite{yang2023dawn}, are grounded in dense architectural paradigms that encounter scalability limitations as data volume and modality complexity increase. To address this, MoE-based frameworks such as CuMo \cite{li2024cumo} and Uni-MoE \cite{Uni_MoE} introduce sparsely-activated expert modules, allowing modality-specific processing while maintaining low inference overhead. CL-MoE \cite{huai2025cl} further extends MoE for continual learning in vision-language tasks, employing dual routers to balance generalization and retention. Furthermore, MoExtend \cite{zhong2024moextend} introduces modular extension mechanisms that facilitate the adaptation of pretrained models to new tasks and modalities, thereby significantly reducing the computational cost associated with full model retraining. 

These approaches illustrate that MoE architectures not only enhance computational efficiency but also offer increased flexibility in handling  multimodal inputs, thereby establishing MoE as a compelling framework for scaling BERT-based models to complex multimodal tasks.
 
\section{Methodology}
\subsection{Overview}
As shown in Figure \ref{fig:MOE}, the MMBERT framework consists of a text tokenizer, word embedding layer, vision and speech encoders, modality aligners, MoE-scaled BERT blocks, and a classification head. Modality aligners project non-text inputs into a shared linguistic space, enabling effective multimodal fusion. The MoE layers are integrated into the BERT encoder to dynamically route representations across modalities, improving detection accuracy. MMBERT is trained in three sequential stages:  Modality aligner training, modality-specific expert training, and MMBERT tuning using a diverse collection of multimodal Chinese hate speech data. The detailed model architecture, training setting and model efficiency information are provided in Appendix A.

\subsection{MMBERT Architecture}
\textbf{Multimodal data generation.} To synthesize the visual and audio data of corresponding text input, we employ the Kokoro text-to-speech model \cite{kaneko2022istftnet} to generate speech data corresponding to the input text. For the visual modality, we render a sequence of word-level font images representing each token in the text, thereby producing a visual analogue of the input. 

\begin{figure*}[t]
  \centering
  \includegraphics[width=\textwidth]{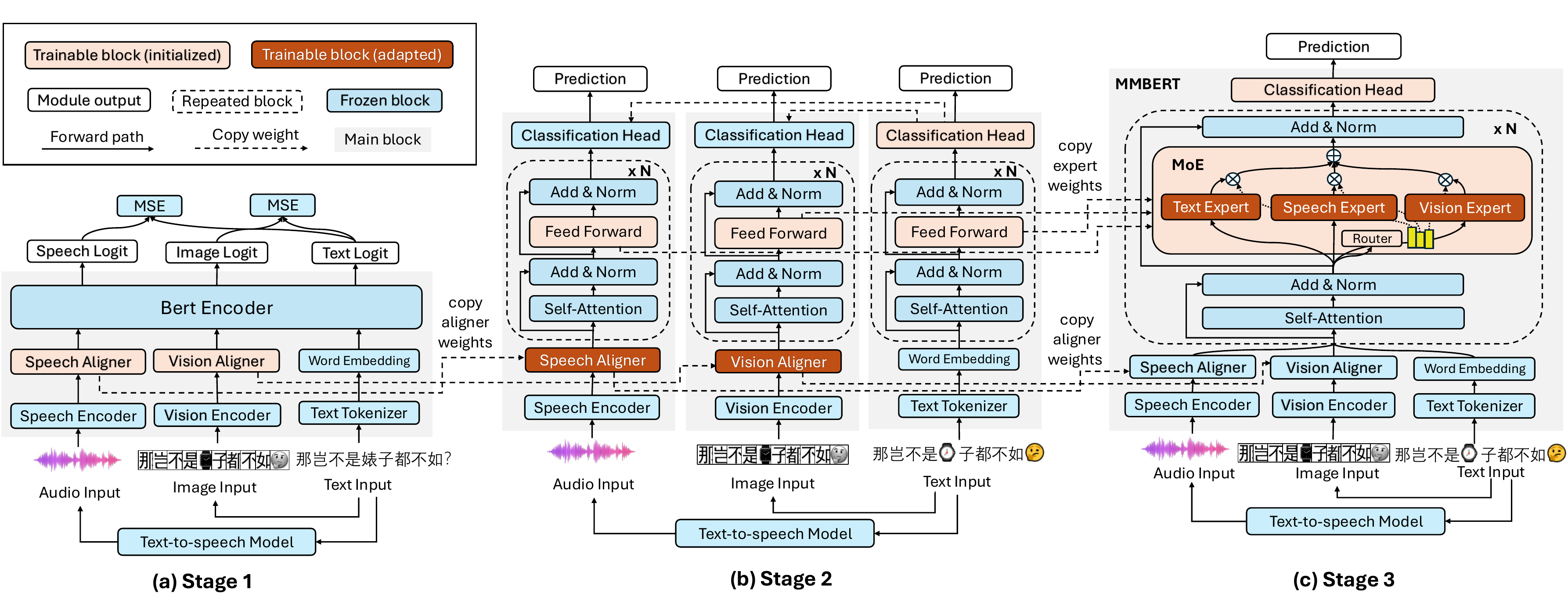}
  \caption{\small \textbf{Illustration of MMBERT Training strategy.} (a) Stage 1: Aligner training, (b) Stage 2: Expert training, (c) Stage 3: MMBERT tuning}
  \label{fig:moe_train}
\end{figure*}

\textbf{Aligners.} To enable the effective transformation of heterogeneous modality inputs into a unified linguistic representation space, MMBERT leverages the pretrained visual-language framework LLaVA \cite{liu2023visual} and the speech-language framework SpeechT5 \cite{ao2021speecht5}. Specifically, for visual encoding, we adopt the CLIP-base-Chinese model \cite{yang2022chinese}, followed by a linear projection layer that maps the extracted visual features into soft image tokens compatible with the embedding space of BERT \cite{devlin2019bert}. For speech, we utilize the encoder from the Whisper-base-Chinese speech recognition model \cite{radford2023robust}, likewise augmented with a linear projection layer to project speech features into the same shared linguistic space. The alignment process is formally defined as follows:
\begin{align}
X &= \{T, \{I_1,\ldots,I_k\}, S\} \\
T &= \text{WordEmbedding}(\text{Tokenizer}(T)) \\
S &= \text{SpeechAligner}(\text{Whisper}(S)) \\
I_i &= \text{VisionAligner}(\text{CLIP}(I_i)) \\
V &= [I_1,\ldots,I_k]
\end{align}
where $\{T, \{I_1,\ldots,I_k\}, S\}$ represents the text, images and speech inputs respectively. The $SpeechAligner$ and $VisionAligner$ modules are implemented as learnable linear projections that transform modality-specific features into a shared language embedding space. The sequence of word-level font image embeddings is concatenated to form the final visual token sequence.

\textbf{MMBERT blocks.} By the above aligners, we could obtain the
encoded embedding of different modalities aligned in unified language domain. We concatenate the different modality embeddings as the final input to the MMBERT blocks.  We denote the text, speech, vision embedding representations to $T = \{T_1,\ldots,T_n\}$, $S = \{S_1,\ldots,S_m\}$ $V = \{V_1,\ldots,V_k\}$ respectively, where 
$n$, $m$, and $k$ correspond to the respective sequence lengths of each modality. The MMBERT block computation proceeds as follows:
\begin{align}
X_{l_0} &= [T_1,\ldots,T_n;\, S_1,\ldots,S_m;\, V_1,\ldots,V_k]\\
X_{l_j}^a &= \text{Self-Atten}(\text{LN}(X_{l_{j-1}})) + X_{l_{j-1}}\\
X_{l_j} &= \text{MoE}(\text{LN}(X_{l_j}^a)) + X_{l_j}^a
\end{align}
where $LN(\cdot)$ refers to layer normalization, the $X_{l_j}^a$ represents the output latent of the self attention layer in the $j$ th MMBERT block, $X_{l_j}$ represents the output latent of $j$ the MMBERT block. The MoE mechanism incorporates a set of experts $E=\{E_T,E_S,E_V\}$ each implemented as a feedforward neural network. A lightweight routing module, implemented as a linear transformation, computes the routing weights that determine the contribution of each modality-specific expert. The process is formally defined as:
\begin{align}
P(X_{l}^a)_i &= \frac{e^{f(X_{l}^a)_i}}{\sum_{m=\{T, S, V\}} e^{f(X_{l}^a)_m}}\\
\text{MoE}(X_l^a) &= \sum_{i=\{T, S, V\}} (P(X_l^a)_i\cdot E_i(X_l^a))
\end{align}
where the $f(\cdot)$ denotes the routing function of different modalities implemented as a linear layer, the output weight logits are normalized by a softmax function. The final MoE output is weighted combination of the different modality-specific expert outputs.

\subsection{MMBERT three-stage training strategy}
To capitalize on the effectiveness of multi-expert collaboration—where each expert possesses distinct capabilities—while retaining the rich contextual and syntactic knowledge encoded in the original BERT model through large-scale pretraining, we propose a three-stage progressive training strategy to facilitate the incremental development of MMBERT. As shown in Figure \ref{fig:moe_train}, the training process is structured into three progressive stages to enhance the efficacy of multi-expert collaboration through an incremental learning strategy.

\textbf{Stage 1: Aligner Training.} The primary objective of the initial stage is to establish effective interoperability between heterogeneous modalities and linguistic representations. Modality-specific MLPs serve as aligners that project inputs from speech and vision into soft token embeddings. These aligners are trained by minimizing the mean squared error between the modality embeddings and the BERT-encoded textual representations. To improve the model's sensitivity to perturbed speech samples, speech and image representations generated from the perturbed text are aligned with those derived from the corresponding unperturbed text representations during the training process.

\textbf{Stage 2: Expert Training.} In this stage, modality-specific experts are trained independently using cross-modal data to specialize in their respective domains. Training continues to be guided by the minimization of cross-entropy loss, while the trained aligners weights in the first stage are adapted and further trained to better capture and represent the unique characteristics inherent to their respective modalities on the Chinese hate speech classification task. To facilitate the projection of heterogeneous modality data into a unified linguistic representation space by both the aligners and experts, the classification head originally trained on textual input is shared across other modalities.

\textbf{Stage 3: MMBERT Tuning.} The final stage integrates the trained experts into the MoE layers of MMBERT. A context-aware routing mechanism dynamically assigns input representations to appropriate experts based on semantic relevance. To prevent unbalanced expert weight distribution, an auxiliary loss is applied to encourage uniform expert utilization:
\begin{align}
\mathcal{L}_{\text{total}} &= \mathcal{L}_{\text{cross-entropy}} + \alpha \cdot \mathcal{L}_{\text{aux}} \\
\mathcal{L}_{\text{aux}} &= N \cdot \sum_{i=1}^{N} p_i \cdot f_i
\end{align}
where $N$ denotes the total number of experts, $\alpha$ represents the weighting coefficient, $p_i$ represents the proportion of sequences routed to expert $i$, and $f_i$ is the average gating probability assigned to expert $i$. The classification head is fine-tuned jointly to generate the final prediction.

\textbf{\begin{table*}[hbtp]
\centering
\sisetup{round-mode=places, round-precision=1, detect-weight=true, detect-mode=true}
\begin{tabular}{@{}l*{12}{S[table-format=2.1]}@{}}
\toprule
\multirow{2.5}{*}{\textbf{Model}} & \multicolumn{4}{c}{\textbf{ToxiCloakCN}} & \multicolumn{4}{c}{\textbf{ToxiCN}} & \multicolumn{4}{c}{\textbf{COLD}} \\
\cmidrule(lr){2-5} \cmidrule(lr){6-9} \cmidrule(lr){10-13}
 & {\textbf{Acc}} & {\textbf{Pre}} & {\textbf{Rec}} & {\textbf{F1}} 
 & {\textbf{Acc}} & {\textbf{Pre}} & {\textbf{Rec}} & {\textbf{F1}} 
 & {\textbf{Acc}} & {\textbf{Pre}} & {\textbf{Rec}} & {\textbf{F1}} \\
\midrule
\multicolumn{13}{c}{\textit{Finetuned Models}} \\
\midrule
LLAMA3-8B   & 78.2 & 79.1 & 77.3 & 79.3 & 81.3 & 82.1 & 83.2 & 84.3 & 78.2 & 78.7 & 80.6 & 78.9 \\
Qwen2.5-7B   & 82.1 & 83.6 & 84.1 & 83.7 & 86.8 & 87.1 & 88.2 & 87.9 & 79.6 & 79.8 & 81.3 & 81.1 \\
BERT          & 80.6 & 80.5 & 80.7  & 86.6 & 87.8 & 88.0 & 87.7& 87.8 & 81.2 & 80.7 & 82.1 & 80.9 \\
BERT-wwm      & 80.0 & 80.4 & 80.3 & 87.9 & 88.0 & 88.1 & 88.9 & 88.0 & 82.0 & 81.6 & 83.2 & 81.8 \\
RoBERTa       & 81.1 & 82.4 & 81.3 & 82.6 & 88.8 & 88.9 & 89.5 & 89.6 & 82.6 & 81.9 & 83.7 & 82.5 \\
ChineseBERT   & 86.3 & 87.5 & 86.2 & 86.8 & 90.8 & 89.4 & 90.3 & 90.6 & 82.4 & 81.3 & 83.1 & 82.2 \\
MMBERT (ours) & \textbf{94.3} & \textbf{94.4} & \textbf{95.7} & \textbf{95.2} & \textbf{93.3} & \textbf{91.4} & \textbf{93.2} & \textbf{92.2} & \textbf{84.2} & \textbf{84.1} & \textbf{86.3} & \textbf{85.8} \\
\bottomrule
\end{tabular}
\caption{Performance comparison of fine-tuned models across datasets with accuracy, precision, recall, and F1 scores.}
\label{tab:model_performance_comparison_tuned}
\end{table*}
}

\section{Experiments}

\subsection{Baseline}
To establish a comprehensive evaluation framework, we consider both encoder-based and decoder-based language models as baselines. Specifically, we adopt several BERT-based models with a fully connected classification layer as encoder-based baselines, and utilize LLMs with structured task-specific prompts as decoder-based baselines.

\textbf{Encoder-Based BERT Models.} As representative encoder-based BERT models, we select three widely adopted Chinese pretrained BERT-based encoders: \textbf{BERT}\footnote{https://huggingface.co/bert-base-chinese} \cite{devlin2019bert}, \textbf{BERT-wwm}\footnote{https://huggingface.co/hfl/chinese-bert-wwm-base} \cite{sun2019ernie} and \textbf{RoBERTa}\footnote{https://huggingface.co/hfl/chinese-roberta-wwm-ext} \cite{liu2019roberta}. Each model is fine-tuned by attaching a fully connected layer on top of the pooled output from the encoder to perform classification. In addition, we include \textbf{ChineseBERT} \cite{sun2021chinesebert}, a recently proposed model that integrates lexicon and phonological features into the standard BERT architecture, to examine its performance under the same experimental settings.

\textbf{Decoder-Based LLMs.} For LLM baselines, we assess the performance of several state-of-the-art LLMs, including \textbf{GPT-3.5} \cite{brown2020language}, \textbf{GPT-4o} \cite{openai2024gpt4o}, \textbf{LLaMA3-8B} \cite{meta2024llama3}, \textbf{Qwen2.5-7B\&72B} \cite{qwen2024qwen2.5}, and \textbf{DeepSeek-v3} \cite{deepseek2024v3}. These models are evaluated under a unified prompt-based inference framework. This setup ensures consistency across different models and enables fair comparison with encoder-based models.

\textbf{\begin{table*}[htbp]
\centering
\sisetup{round-mode=places, round-precision=1, detect-weight=true, detect-mode=true}
\begin{tabular}{@{}l*{12}{S[table-format=2.1]}@{}}
\toprule
\multirow{2.5}{*}{\textbf{Model}} & \multicolumn{4}{c}{\textbf{ToxiCloakCN}} & \multicolumn{4}{c}{\textbf{ToxiCN}} & \multicolumn{4}{c}{\textbf{COLD}} \\
\cmidrule(lr){2-5} \cmidrule(lr){6-9} \cmidrule(lr){10-13}
 & {\textbf{Acc}} & {\textbf{Pre}} & {\textbf{Rec}} & {\textbf{F1}} 
 & {\textbf{Acc}} & {\textbf{Pre}} & {\textbf{Rec}} & {\textbf{F1}} 
 & {\textbf{Acc}} & {\textbf{Pre}} & {\textbf{Rec}} & {\textbf{F1}} \\
\midrule
\multicolumn{13}{c}{\textit{LLM APIs (2 unperturbed hate / non-hate speech examples)}} \\
\midrule
GPT-3.5        & 55.5 & 60.5 & 55.5 & 49.5 & 60.7 & 63.7 & 60.7 & 58.5 & 65.2 & 73.6 & 64.9  & 61.3 \\
GPT-4o        & 64.5 & 68.8 & 64.6 & 62.4 & 76.2 & 76.8 & 76.3 & 76.4 & 71.5 & 73.4 & 71.5 & 70.9 \\
LLAMA3-8B    & \textbf{68.2} & 68.2 & \textbf{68.1} & 68.0 & 74.2 & 74.2 & 74.1 & 74.1 & 70.6 & 70.8 & 70.6 & 70.6 \\
Qwen2.5-7B    & 66.0 & 66.7 & 66.0 & 65.6 & 76.4 & 77.3 & 76.4 & 76.2 & \textbf{74.7} & 76.1 & 74.7 & 74.3 \\
DeepSeek-v3   & 64.6 & 68.3 & 64.5 & 66.2 & 72.9 & 77.5 & 72.8 & 71.7 & 73.1 & 75.4 & 73.1 & 72.5 \\
Qwen2.5-72B   & 67.9 & \textbf{69.2} & 67.2 & \textbf{68.1} & \textbf{77.3} & \textbf{78.6} & \textbf{77.1} & \textbf{77.9} & 74.6 & \textbf{77.1} & \textbf{75.3} & \textbf{74.7} \\
\midrule
\multicolumn{13}{c}{\textit{LLM APIs ((2 unperturbed \& 2 perturbed hate / non-hate examples)}} \\
\midrule
GPT-3.5        & 55.3 & 61.2 & 55.7 & 49.8 & 60.3 & 63.5 & 61.2 & 58.2 & 65.4 & 73.7 & 65.1 & 61.4 \\
GPT-4o         & 66.9 & 71.2 & 68.3 & 67.8 & 78.1 & \textbf{79.9} & 78.1 & 77.8 & 71.5 & 73.4 & 71.5 & 70.9 \\
LLAMA3-8B     & 67.3 & 68.9 & 67.9 & 68.2 & 75.1 & 74.0 & 74.2 & 74.3 & 71.2 & 70.7 & 72.1 & 71.2 \\
Qwen2.5-7B    & 65.9 & 66.5 & 66.4 & 66.1 & 77.2 & 78.6 & 77.2 & 77.1 & 75.2 & 76.3 & 74.7 & 75.8 \\
DeepSeek-v3   & 68.2 & \textbf{70.2} & 67.1 & 65.2 & 73.8 & 77.1 & 74.3 & 73.7 & 75.9 & \textbf{77.6} & 74.2 & 75.3 \\
Qwen2.5-72B   & \textbf{71.2} & 69.7 & \textbf{71.1} & \textbf{68.3} & \textbf{78.4} & 79.3 & \textbf{78.2} & \textbf{78.6} & \textbf{76.9} & 76.9 & \textbf{76.2} & \textbf{76.1} \\
\midrule
\multicolumn{13}{c}{\textit{LLM APIs (2 unperturbed \& 2 perturbed hate / non-hate examples \& CoT )}} \\
\midrule
GPT-3.5        & 57.3 & 62.3 & 58.1 & 51.6 & 62.9 & 65.8 & 61.2 & 59.3 & 66.1 & 73.8 & 63.2 & 63.4 \\
GPT-4o        & 71.5 & 72.1 & 67.6 & 69.3 & 79.4 & 81.2 & 79.9 & 79.8 & 74.2 & 76.4 & 74.3 & 73.8 \\
LLAMA3-8B     & 70.1 & 69.2 & 66.4 & 68.2 & 76.4 & 73.8 & 75.2 & 74.8 & 71.4 & 70.3 & 70.8 & 70.7 \\
Qwen2.5-7B    & 68.1 & 67.1 & 65.8 & 66.1 & 77.4 & 76.9 & 77.8 & 77.3 & 75.1 & 75.9 & 75.8 & 74.9 \\
DeepSeek-v3   & 70.6 & \textbf{72.4} & 72.5 & \textbf{71.6} & 76.6 & \textbf{81.5} & 78.3 & 77.1 & 78.2 & \textbf{81.3} & 76.9 & 77.3 \\
Qwen2.5-72B   & \textbf{72.3} & 71.8 & \textbf{72.7} & 70.3 & \textbf{81.1} & 80.7 & \textbf{81.3} & \textbf{80.1} & \textbf{78.4} & 78.5 & \textbf{78.1} & \textbf{78.2} \\
\bottomrule
\end{tabular}
\caption{Performance comparison of LLM prompting across datasets with accuracy, precision, recall, and F1 scores.}
\label{tab:model_performance_comparison_api}
\end{table*}
}

\subsection{Dataset} 

To evaluate the proposed MMBERT, we conduct experiments on three Chinese hate speech datasets that collectively support comprehensive and robust assessment. \textbf{ToxiCN} \cite{lu2023facilitating} provides 12,011 samples of standard hate speech annotations for naturally occurring Chinese text, serving as a baseline for evaluating classification performance. \textbf{ToxiCloakCN} \cite{xiao2024toxicloakcn} introduces 4,582 cloaking perturbed examples in code-mixing and homophonic substitution, specifically designed to evade text-only detectors while preserving hateful intent, making it essential for testing model robustness against cloaking strategies. Finally, \textbf{COLD} \cite{deng2022coldbenchmarkchineseoffensive} extends evaluation to a wider spectrum of offensive content with 37,480 samples, offering insight into a model’s generalizability across various forms of toxicity. Together, these datasets form a diverse and challenging benchmark suite for assessing both accuracy and adversarial resilience in Chinese hate speech detection.

\begin{figure*}[t]
    \centering    \includegraphics[width=\textwidth]{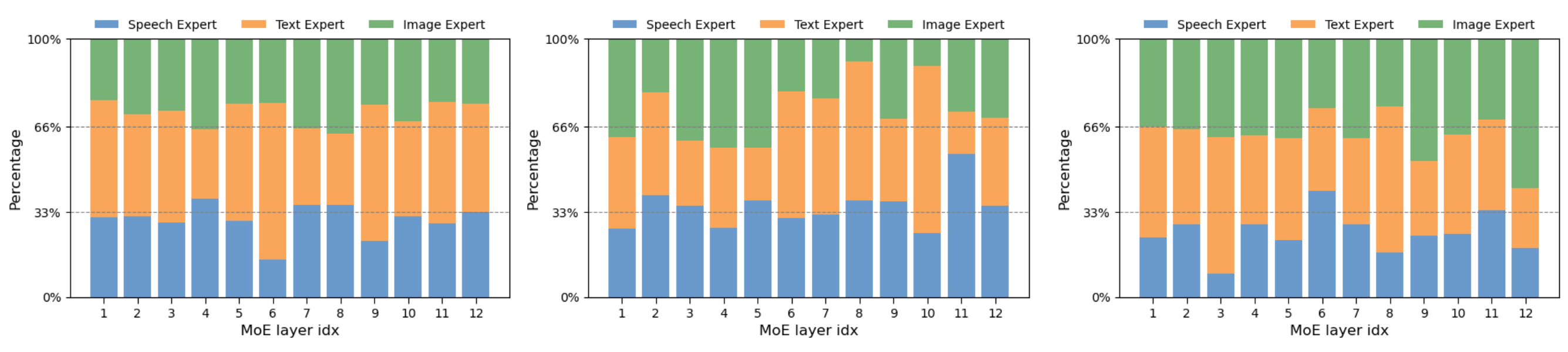}
    \caption{\small Distribution of expert loading with different input perturbation types, \textit{left}: non perturbation, \textit{middle}: homophonic perturbation, \textit{right}: code-mixing perturbation} 
    \label{fig:moe_dist}
\end{figure*}

\subsection{Evaluation method}

We employ the widely used metrics of accuracy (\textbf{Acc}), macro precision (\textbf{Pre}), macro recall (\textbf{Rec}) and macro $F_1$-score (\textbf{F1}) to evaluate the classification performance of models. For the BERT-based models and open source LLMs with relatively comparable parameter size with MMBERT in the baselines, we fine-tune and reserve the best performing models with hyperparameters on the test set. All datasets are partitioned into training, validation and test sets using an 8:1:1 split ratio with early stopping strategy to prevent overfit during training. For the LLMs in the baselines, we perform few-shot learning with a basic prompt temple with different few-shot learning and chain‑of‑thought (CoT) settings, details can be found in Appendix B. All experiments are conducted using a NVIDIA H100 Tensor Core GPU.

\subsection{Result and Discussion}
\subsubsection{Main result}

Table~\ref{tab:model_performance_comparison_tuned} and~\ref{tab:model_performance_comparison_api} presents the evaluation of fine‑tuned LLMs, BERT‑based models and LLM APIs across the ToxiCloakCN, ToxiCN, and COLD benchmarks using accuracy, macro precision, macro recall, and macro F1 as metrics. MMBERT consistently achieves the highest scores across all three datasets, demonstrating both strong overall performance and robustness to adversarial perturbations. Specifically, MMBERT attains macro F1 scores of 95.2, 92.2, and 85.8 on ToxiCloakCN, ToxiCN, and COLD, respectively. Compared to the strongest fine‑tuned baseline, ChineseBERT, these results represent improvements of 8.4, 1.6, and 3.6 points in macro F1. These gains highlight the effectiveness of integrating textual, speech, and visual modalities through the Mixture‑of‑Experts framework and the progressive three‑stage training strategy, which jointly enhance the model’s ability to capture phonological and visual cues indicative of cloaked hate speech.  

Traditional encoder‑based models, including BERT, RoBERTa, and ChineseBERT, perform competitively on ToxiCN and moderately well on COLD. However, their performance drops substantially on ToxiCloakCN, confirming their vulnerability to character deformation, homophonic substitution, and code‑mixing perturbations. In contrast, LLM APIs such as GPT‑3.5, GPT‑4o, LLaMA3‑8B, Qwen2.5‑7B, and DeepSeek‑v3 show limited effectiveness in few‑shot and perturbed settings. For example, GPT‑4o achieves only 62.4~F1 on ToxiCloakCN under basic prompting, underscoring the insufficiency of in‑context learning alone for this domain‑specific and adversarial task.  

Providing both unperturbed and perturbed examples, as well as incorporating CoT prompting, yields modest improvements for LLMs. GPT‑4o, for instance, improves from 62.4 to 69.3~F1 on ToxiCloakCN under the CoT setting. Nevertheless, these enhancements remain far below the performance of MMBERT, indicating that domain‑adaptive multimodal modeling is critical for robust detection rather than relying solely on prompting.  

Across datasets, ToxiCloakCN poses the greatest challenge due to heavy use of cloaking perturbations, and MMBERT is the only model to surpass 90~F1 on this benchmark. ToxiCN represents standard hate speech detection, where all fine‑tuned BERT variants perform strongly and MMBERT provides consistent incremental gains. COLD, as a more diverse and open‑domain dataset, produces lower overall scores, yet MMBERT maintains the best recall, confirming its generalization to nuanced and implicit toxic language.  

Overall, the results validate the task-specific multimodal modeling with MoE-based expert routing and progressive training for MMBERT substantially outperforms both fine-tuned text-only models and prompt-based LLMs, particularly in adversarial scenarios involving cloaked hate speech. Detailed failure case analyses are presented in Appendix C.

\subsubsection{Routing distribution analysis} We analyze the average routing weight distribution of different experts in MMBERT 12 MoE layers under three hate speech perturbation categories in the ToxiCloakCN dataset as shown in Figure \ref{fig:moe_dist}. 

In the non-perturbed setting, the model primarily routes to the text expert, especially in middle layers, reflecting the dominance of textual semantics. Speech and image experts contribute consistently, with image usage slightly increasing in deeper layers. Under homophonic perturbation, the model shifts toward the speech expert in early and middle layers, leveraging phonetic cues to resolve ambiguities introduced by homophones. Vision expert assigned weight decreases slightly, while text routing remains stable. In the code-mixing scenario, image experts dominate across most layers, indicating reliance on visual context to address multilingual inconsistencies. Text experts are also more engaged in earlier layers, while speech expert weight declines. 

These patterns demonstrate MMBERT adaptive routing behavior, where expert activation is dynamically adjusted based on input characteristics, enhancing robustness against modality-specific perturbations.

\subsubsection{Ablation study on training strategy} We conduct an ablation study to evaluate the effectiveness of the progressive three-stage training strategy for integrating MoE into MMBERT. Specifically, we compare the full pipeline with three variants: without aligner training stage (stage 1),  without expert training stage (stage 2), and without both stages. All models are trained for 50 epochs on the ToxiCloakCN dataset under identical settings.

\begin{figure}[t]
    \centering
\includegraphics[width=0.5\textwidth,height=0.32\textwidth]{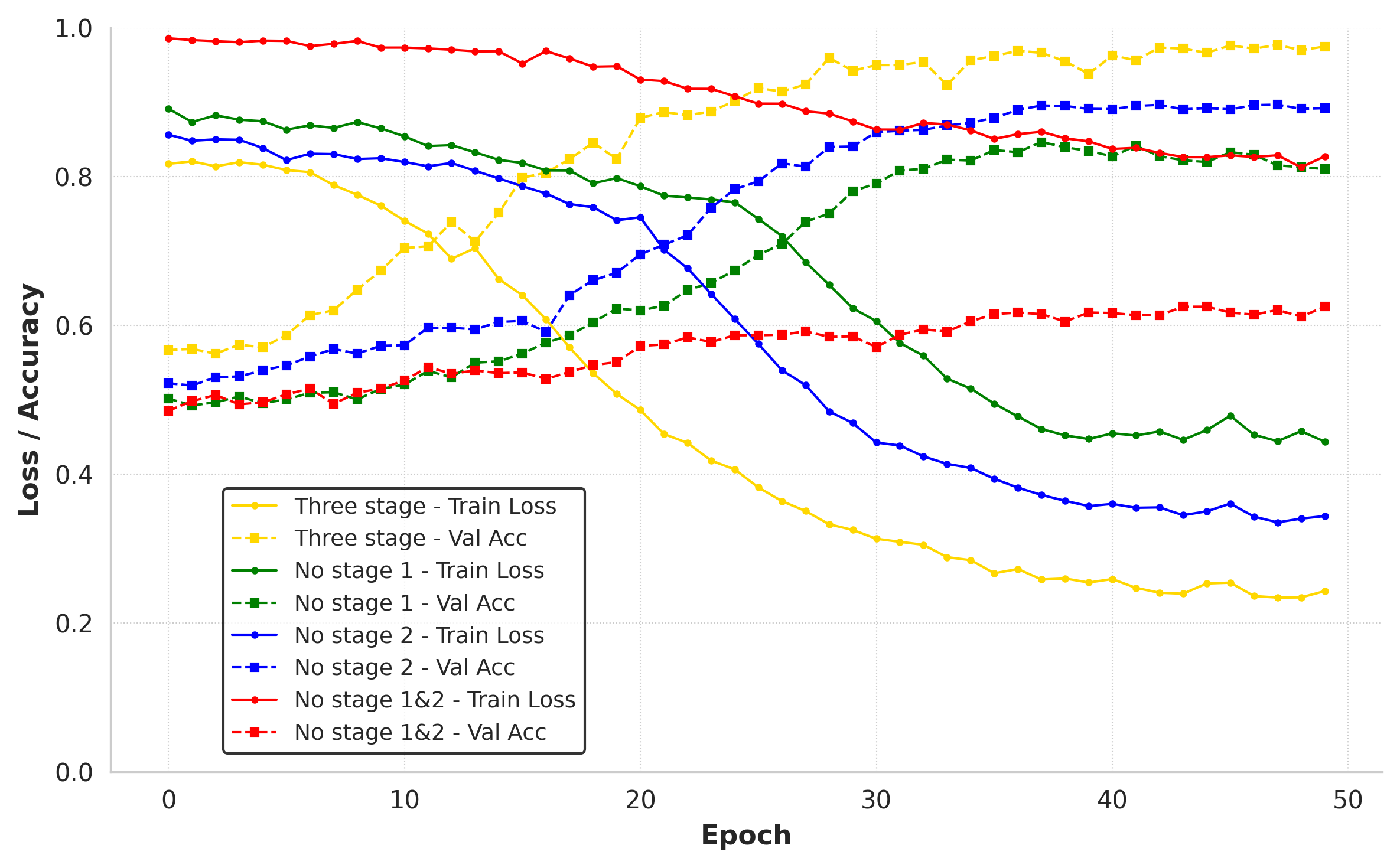}
    \caption{\small Ablation study evaluating the impact of each stage in the proposed three-stage training strategy}
    \label{fig:training_ablation}
\end{figure}

As shown in Figure~\ref{fig:training_ablation}, the full three-stage strategy achieves the best overall performance, with the lowest training loss and highest validation accuracy. It enables stable convergence and strong generalization, indicating that gradual modality alignment and expert specialization are both essential for effective multimodal learning. Without aligner pretraining, convergence is slower and validation performance is less stable, suggesting suboptimal cross-modal mapping. Removing expert specialization also leads to reduced accuracy and higher loss, showing that expert-specific representation learning is crucial. The worst performance is observed when both stages are removed, as the model quickly overfits and fails to generalize. These results demonstrate that each stage of the proposed training strategy plays a critical role in enabling MMBERT to effectively detect cloaked hate speech across modalities.

\begin{table}[t]
\centering
\sisetup{round-mode=places, round-precision=1, detect-weight=true, detect-mode=true}
\begin{tabular}{@{}l*{12}{S[table-format=2.1]}@{}}
\toprule
\multirow{2.5}{*}{\textbf{Dataset}} & \multicolumn{2}{c}{\textbf{Text\&Speech}} & \multicolumn{2}{c}{\textbf{Text\&Vision}} \\
\cmidrule(lr){2-3} \cmidrule(lr){4-5} 
 & {\textbf{Acc}} & {\textbf{F1}} 
 & {\textbf{Acc}} & {\textbf{F1}} \\
\midrule
ToxiCloakCN   & 91.2 & 91.1 & 87.7  & 86.6   \\
ToxiCN        & 90.1 & 90.9 & 88.9  & 89.3   \\
COLD          & 83.1 & 83.8 & 82.7  & 81.9   \\
\bottomrule
\end{tabular}
\caption{Ablation study evaluating the impact of each modality in the MMBERT framework}
\label{tab:modality_ablation}
\end{table}

\subsubsection{Ablation study on modalities} To assess the contribution of each modality in the MMBERT framework, we perform an ablation study by scaling with single modality, using text paired with either speech or vision. As shown in Table \ref{tab:modality_ablation}, the text and speech combination consistently outperforms the text and vision setting across all three datasets. On the ToxiCloakCN dataset, the F1 score reaches 91.1 when using speech compared to 86.6 when using vision, indicating that speech features are more effective in capturing adversarial cues introduced by cloaking perturbations. This trend is also observed on ToxiCN and COLD, where the text and speech setting yields stronger results. These findings suggest that speech contributes more complementary information than vision and plays a critical role in improving robustness in Chinese hate speech detection.

\section{Conclusion}
We presents MMBERT, a multimodal framework for Chinese hate speech detection that effectively incorporates text, speech, and vision using the MoE architecture. To ensure stable integration of modalities, we introduce a progressive training strategy that proves critical for effective optimization. Ablation studies confirm the importance of both the training strategy and modality fusion, with speech contributing significantly to robustness. Empirical results across multiple benchmarks show that MMBERT achieves strong performance, particularly under adversarial conditions involving cloaked perturbations. Our findings highlight the potential of task-specific multimodal modeling for addressing complex language understanding challenges, particularly in safety-critical domains like Chinese hate speech detection.

\section{Ethics Statement}

This work involves Chinese hate speech detection with sensitive content. 
All datasets are publicly available and anonymized, and our models are intended solely for research to avoid potential bias and misuse.


\newpage

\appendix
\section*{Appendix A: MMBERT Details}
\label{sec:appendix_mmbert_setting}
\subsection{Model Architecture} 
MMBERT is built upon the \texttt{BERT-base-chinese}\footnote{https://huggingface.co/bert-base-chinese} encoder, which serves as the backbone for textual representation. For modality-specific feature extraction, we employ a vision encoder based on \texttt{chinese-clip-vit-base-patch16}\footnote{https://huggingface.co/OFA-Sys/chinese-clip-vit-base-patch16} and a speech encoder based on \texttt{whisper-base}\footnote{https://huggingface.co/openai/whisper-base}. Each modality is passed through a dedicated aligner, implemented as a lightweight two-layer MLP, to project the modality-specific features into the BERT embedding space, thereby forming unified token representations. These representations are processed by modified BERT layers in which the original feed-forward networks are replaced by Mixture-of-Experts (MoE) layers. Each MoE layer contains modality-specific experts and a shared self-attention mechanism, with a context-aware routing function that dynamically assigns token sequences to appropriate experts. A classification head is applied to the final output to produce predictions.

\subsection{Training Setting}
Training is performed in three progressive stages. In stage 1, modality aligners are pretrained using synthetic parallel data to align visual and speech features with their corresponding textual embeddings. The learning rate in this stage is set to 1e-3. In stage 2, modality-specific experts are trained independently using cross-modal supervision, while aligners continue to adapt. During this phase, the learning rate for the aligners is maintained at 1e-3, the text expert at 5e-6, and the speech and vision experts at 5e-5. In stage 3, all components are jointly fine-tuned on the multimodal Chinese hate speech detection task using a cross-entropy loss. The learning rate in this final stage is set to 5e-4. To promote balanced utilization across experts, we incorporate an auxiliary load-balancing loss into the MoE layers, with a weighting coefficient of 1e-2. 

The model is trained for 50 epochs using the AdamW optimizer and a linear learning rate decay schedule. Excluding the parameters of the modality-specific encoders, the MMBERT architecture contains approximately 60 million trainable parameters. All experiments are conducted using PyTorch on NVIDIA A100 GPUs. 

\subsection{Model Efficiency}
\textbf{Parameter Count.}
The MMBERT model comprises 297.4 million parameters in total, including 162.4M in the backbone network (representing a 47\% increase relative to BERT‑base), 49M in the Whisper‑base speech encoder, and 86M in the CLIP‑base vision encoder.

\textbf{Computational Cost.}
A single forward pass requires approximately 58.44 GFLOPs, which is the sum of 12×2.89 GFLOPs from the MMBERT layers, 21.2 GFLOPs from the Whisper‑base encoder, and 2.56 GFLOPs from the CLIP‑base encoder. The contribution of the pooler and classifier heads is negligible.

\textbf{Routing Overhead.}
The mixture‑of‑experts (MoE) layer routing introduces an additional 908.4 MFLOPs (12×75.8 MFLOPs), accounting for approximately 2.6\% of the total computational cost.

\textbf{Inference Latency.}
Under single‑query inference with a sequence length of 128 on an NVIDIA H100 GPU, MMBERT achieves a latency of 6.3 ms in FP32 precision (compared to 3.5 ms for BERT‑base) and 3.2 ms in FP16 precision (compared to 2 ms for BERT‑base).

\section{Appendix B: LLM evaluation prompt template}
\label{sec:llm_template}
\begin{figure}[H]
    \centering
    \includegraphics[width=\columnwidth]{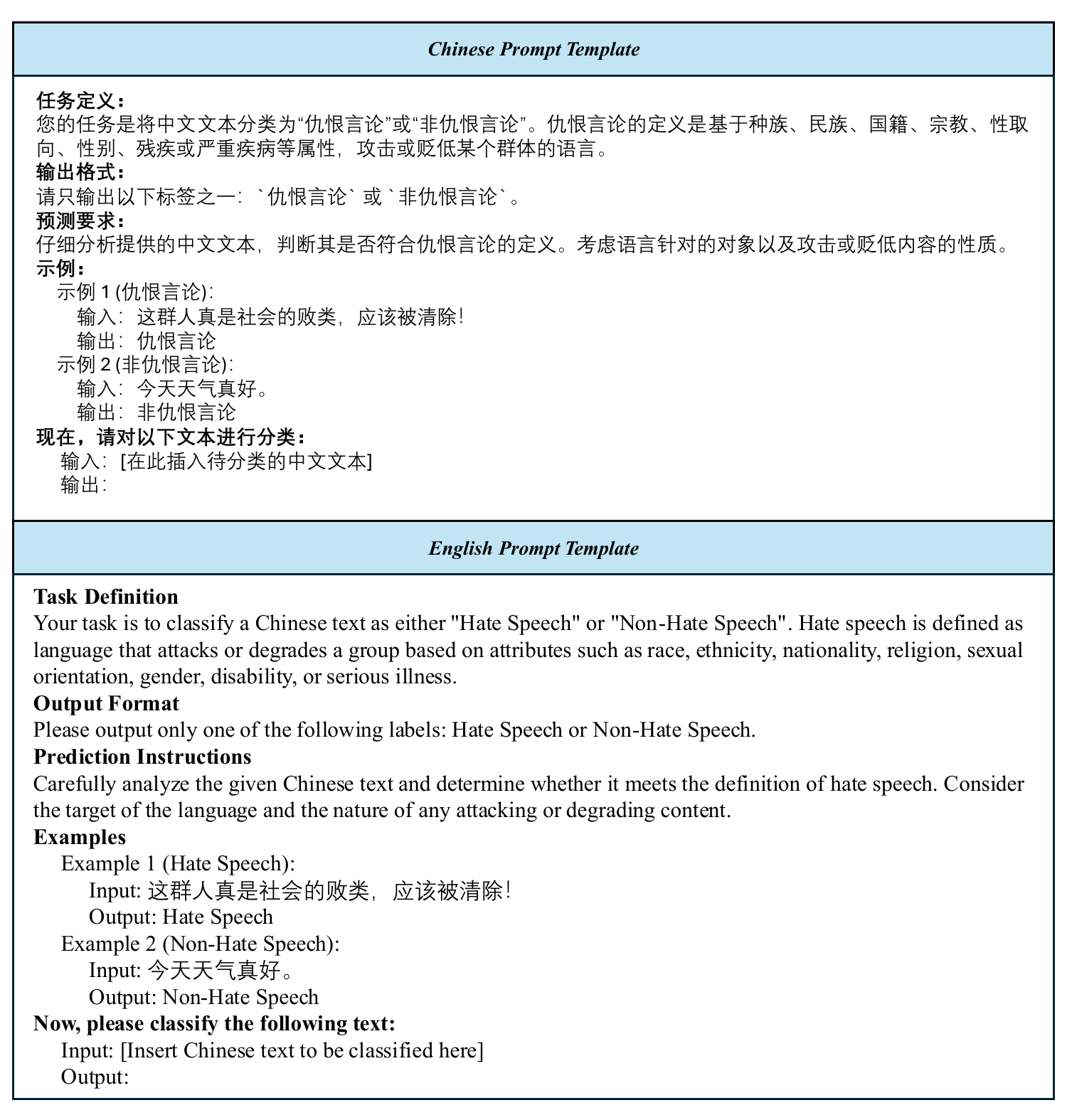}
    \caption{\small Chinese and English version of the LLM Chinese hate speech detection evaluation template}
    \label{fig:llm_template}
\end{figure}

The basic prompt template structure of LLM prompting for Chinese hate speech detection is shown in Figure \ref{fig:llm_template}

\section*{Appendix C: Failure Case Analysis}

To better understand the limitations of MMBERT, 
we manually reviewed 50 misclassified samples from each test set. 
Two dominant failure modes emerged:

\subsection*{Cultural Context Gaps (38\%)}
\textbf{False Positive Example (COLD):}  
\begin{quote}
\textit{``Taiwanese rednecks leave Weibo''}
\end{quote}
\textbf{Root Cause:}  
The model misclassifies culturally nuanced expressions as toxic due to 
limited coverage of regional dialects and sociopolitical context in the training data.  

\textbf{Mitigation Strategy:}  
Diversify annotation teams with native speakers from multiple 
Chinese-speaking regions and include context-rich examples to reduce such errors.

\subsection*{Sarcasm and Reclaimed Terms (32\%)}
\textbf{True Negative Example (ToxiCN):}  
\begin{quote}
\textit{``We gays are disgusting haha''}
\end{quote}
\textbf{Root Cause:}  
Binary toxicity labels lack contextual nuance. The model cannot 
distinguish reclaimed slurs or self-deprecating humor from genuine hate.  

\textbf{Mitigation Strategy:}  
Introduce ternary labeling schemes (e.g., \textit{hate}, \textit{reclaimed}, \textit{neutral}) 
or enrich the dataset with metadata such as speaker identity and intent.
\subsection{}
These errors highlight that MMBERT is sensitive to 
cultural variation, sarcasm, and reclaimed language. 
Future work should explore context-aware annotations, 
richer label taxonomies, and sociolinguistic metadata to 
improve robustness in real-world deployment.


\end{document}